\documentclass[]{llncs}
\usepackage{graphicx}
\usepackage{amsmath}
\usepackage{cite}
\usepackage{algorithmic}
\usepackage{algorithm}
\usepackage{array}
\usepackage{textcomp}
\usepackage{stfloats}
\usepackage{url}
\usepackage{verbatim}
\usepackage{color}
\usepackage{booktabs}
\usepackage{xcolor}
\usepackage{multirow}
\usepackage{enumitem}
\definecolor{citecolor}{HTML}{0071bc} 
\usepackage[colorlinks, linkcolor=red,  anchorcolor=blue, citecolor=citecolor]{hyperref}
\usepackage{float}
\definecolor{SeaGreen4}{RGB}{0,205,102} 
\definecolor{SlateBlue}{RGB}{106,90,205} 
\definecolor{DarkRed}{RGB}{178,34,34} 
\usepackage{times}
\usepackage{soul}
\usepackage[switch]{lineno}
\usepackage{amssymb}
\usepackage{pifont}
\usepackage{makecell}
\usepackage{colortbl}
\definecolor{mygray}{gray}{.9}
\definecolor{mypink}{rgb}{.99,.91,.95}
\definecolor{mycyan}{cmyk}{.3,0,0,0}
\usepackage[misc]{ifsym}

\usepackage{subcaption}  

\begin{document}

\title{EMDFNet: Efficient Multi-scale and Diverse Feature Network for Traffic Sign Detection
}  


\author{Pengyu Li, Chenhe Liu, Tengfei Li, Xinyu Wang, Shihui Zhang, Dongyang Yu} 



\author{Pengyu Li\inst{1} \and
Chenhe Liu\inst{1} \and 
Tengfei Li\inst{1} \and 
Xinyu Wang\inst{1} \and \\
Shihui Zhang\inst{1} \thanks{Corresponding author: Shihui Zhang, Email: (\email{sshhzz@ysu.edu.cn}) \\ 
The work is partially supported by the Hebei Natural Science Foundation No.F2023203012.} \and
Dongyang Yu\inst{2}
}

\authorrunning{Pengyu Li et al.}


\institute{ Yanshan University, Qinhuangdao City, 066000, Hebei Province, China
\email{\{202111130221, 202211200217, funlee, zora030\}@stumail.ysu.edu.cn, sshhzz@ysu.edu.cn}\\
\and
Beijing Rigour Technology Co.,Ltd.\\
\email{yudongyang2022@gmail.com}}

\maketitle           

\begin{abstract}
The detection of small objects, particularly traffic signs, is a critical subtask within object detection and autonomous driving. Despite the notable achievements in previous research, two primary challenges persist. Firstly, the main issue is the singleness of feature extraction. Secondly, the detection process fails to effectively integrate with objects of varying sizes or scales. These issues are also prevalent in generic object detection. Motivated by these challenges, in this paper, we propose a novel object detection network named Efficient Multi-scale  and Diverse Feature  Network (EMDFNet) for traffic sign detection that integrates an Augmented Shortcut Module and an Efficient Hybrid Encoder to address the aforementioned issues simultaneously. Specifically, the Augmented Shortcut Module utilizes multiple branches to integrate various spatial semantic information and channel semantic information, thereby enhancing feature diversity. The Efficient Hybrid Encoder utilizes global feature fusion and local feature interaction based on various features to generate distinctive classification features by integrating feature information in an adaptable manner. Extensive experiments on the Tsinghua-Tencent 100K (TT100K) benchmark and the German Traffic Sign Detection Benchmark (GTSDB) demonstrate that our EMDFNet outperforms other state-of-the-art detectors in performance while retaining the real-time processing capabilities of single-stage models. This substantiates the effectiveness of EMDFNet in detecting small traffic signs.
\end{abstract}

\section{Introduction}  
Traffic sign detection and recognition (TSD and TSR) are crucial in intelligent transportation and autonomous driving. They aim to detect traffic signs and identify their types for information extraction and safety enhancement. Unlike some studies that separate "detection" (localization) and "recognition" (classification), we consider that TSD includes both in a broader range of object detection.

\begin{figure}[!t]
	\centering
	\includegraphics[scale=0.15]{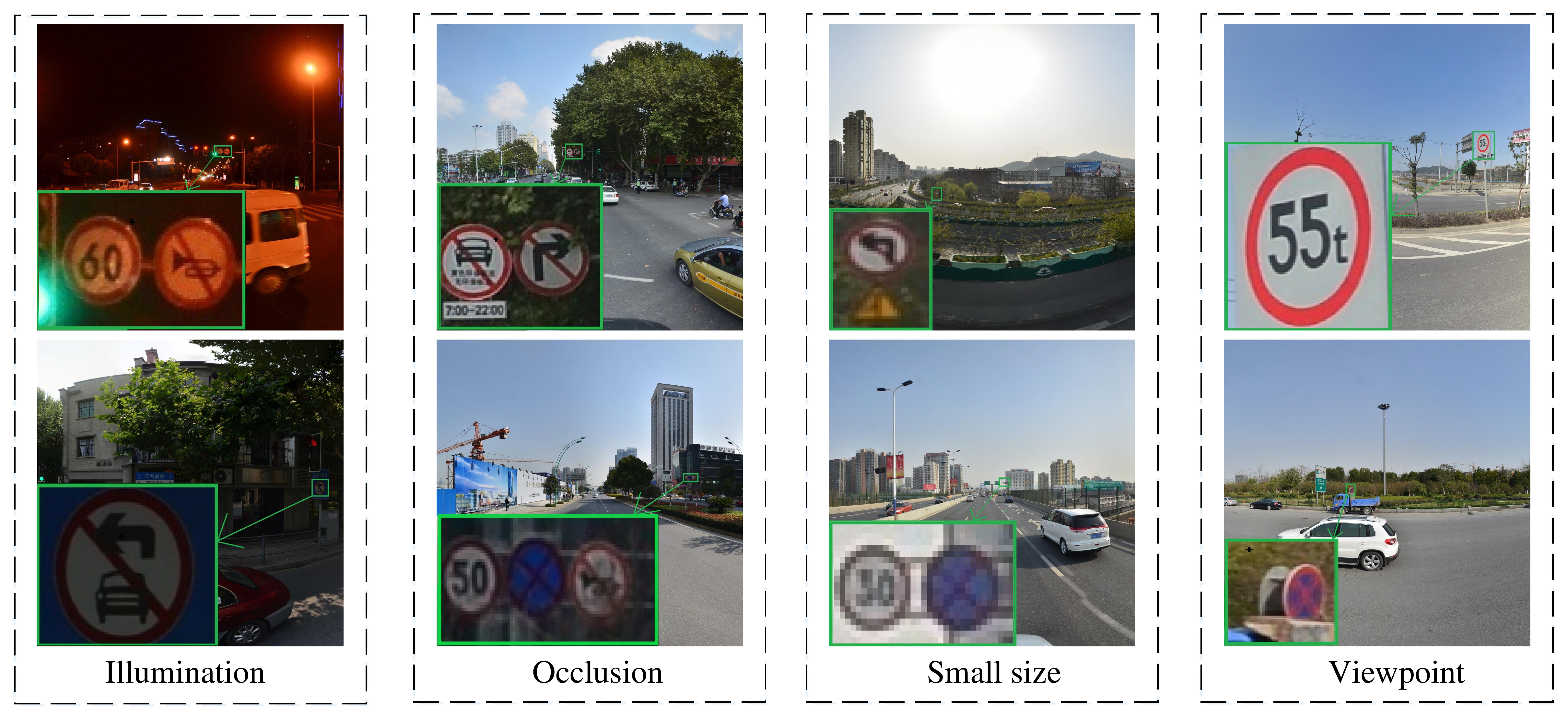}
	\caption{The difficulties in traffic sign detection. In real traffic scenes, traffic sign detection faces many difficulties including illumination, occlusion, small size, viewpoint and so on.}
\label{challenge dataset}
\vspace{-0.3cm}
\end{figure}

TSD faces challenges, notably inefficiency due to small sign sizes at long distances, exemplified by signs covering only about $25 \times 25$ pixels in $2000 \times 2000$ pixel images. The issue of size, combined with the tendency of generic detection models to focus on smaller input resolutions, increases the risk of missing small signs. Additionally, as shown in Fig.~\ref{challenge dataset}, detection accuracy is compromised by factors such as complex background, occlusion, deformation, and change in lighting.

Motivated by the above problems, we propose an Efficient Multi-scale  and Diverse Feature  Network for traffic sign detection in complex background. The contributions of this paper are summerized as four-folds:

$\bullet$ We propose an Efficient Multi-scale  and Diverse Feature  Network (EMDFNet) for small object detection in complex background. The network possesses a highly powerful capability for feature extraction and multi-scale interaction, thereby enhancing the robustness of small object detection. The network achieves good trade-offs in terms of completeness, real-time processing, and accuracy.

$\bullet$ We propose an Augmented Shortcut Module (ASM) to address the issue of feature singularity. The module combines different spatial semantic information and channel semantic information through multiple branches to enhance feature diversity. Our ASM excels at capturing the fundamental visual patterns and semantic information of small objects, thereby enhancing the effectiveness of small object detection.

$\bullet$ To enhance multi-scale detection capability of the network, we introduce the Efficient Hybrid Encoder (EHE) to improve the neck structure. This encoder combines intra-scale feature interaction and cross-scale feature fusion. To the best of our knowledge, we are the first to apply the idea of EHE to the field of traffic sign detection.

$\bullet$ Extensive experiments are conducted on two widely used datasets TT100K and GTSDB. The experimental results in both qualitative and quantitative dimensions demonstrate that the proposed EMDFNet reliably achieves state-of-the-art performance in traffic sign detection.

\section{Related Work} 

\subsection{Object Detection} 

Deep convolutional neural networks (CNNs) have become the preferred choice for various computer vision tasks in recent years. They have significantly improved the accuracy and speed of object detection, as evidenced in benchmarks such as Pascal VOC\cite{everingham2010pascal} and MS COCO\cite{lin2014microsoft}. The R-CNN series, including R-CNN\cite{girshick2014rich}, Fast R-CNN\cite{girshick2015fast}, Faster R-CNN\cite{ren2015faster}, and Mask R-CNN\cite{he2017mask}, extract region proposals before classifying and regressing them for detection results. However, their two-stage process is computationally intensive, limiting their practical use. Single-stage detectors like SSD\cite{liu2016ssd} and YOLO eliminate the proposal step to enable real-time performance but face challenges with small objects because of the loss of spatial and channel details. This paper aims to preserve spatial and channel feature information to enhance the detection performance of small traffic signs.

\subsection{Traffic Sign Detection} 
Traffic sign detection (TSD) is a specialized area within object detection, where generic methods are less effective due to the small size and complex detection requirements of traffic signs. Traditional techniques focus on handcrafted features such as color, shape, and edges, often combined with machine learning classifiers. However, these methods struggle with environmental variations such as lighting and motion blur.

Recent advancements have seen the rise of CNN-based approaches, utilizing strategies such as fully convolutional networks (FCN)\cite{zhu2016trafficonv} for sign proposals and deep CNNs for classification. Additionally, techniques involving image pyramids and Generative Adversarial Networks (GANs) have been employed to enhance detection performance, particularly for small objects. Despite successes, limitations remain in data availability and sign size, impacting the balance between speed, accuracy, and detection completeness. This paper focuses on the trade-off between accuracy and speed.

\begin{figure*}[!t]
	\centering
	\includegraphics[scale=0.42]{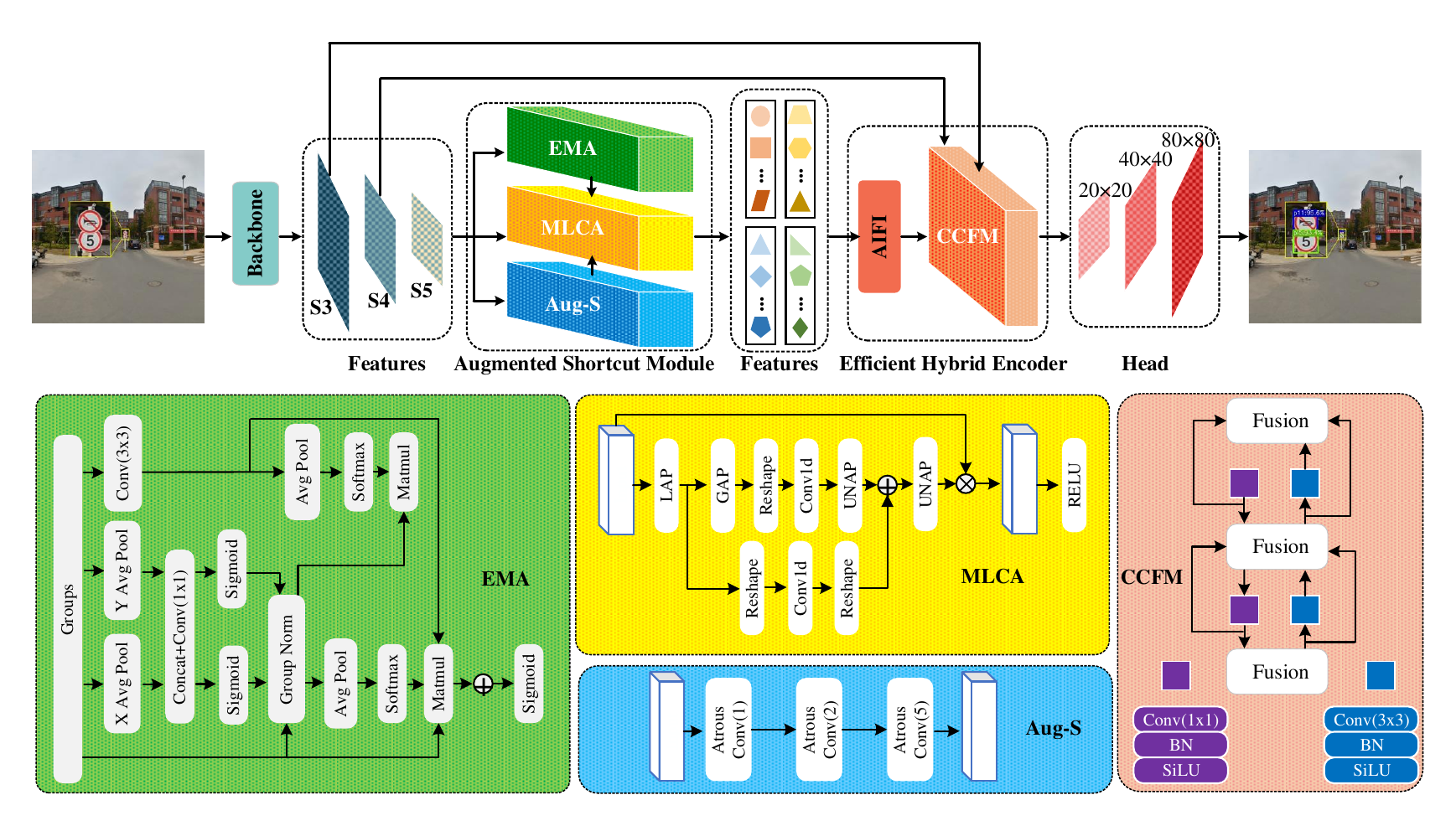}
	\caption{The EMDFNet can be roughly divided into four parts. The first part consists of a backbone composed of Res2Net. The second part is the Augmented Shortcut Module, which enhances feature diversity. The third part is the Efficient Hybrid Encoder for integrating multi-scale features. Finally, the fourth part comprises three prediction heads for bounding box prediction.}
\label{figure2}
\end{figure*}


\section{Proposed Network} 

\subsection{Network Overview} 
As illustrated in Fig.~\ref{figure2}, our EMDFNet, an end-to-end object detection network that incorporates multi-scale feature fusion and diverse features. EMDFNet leverages Res2Net \cite{gao2019res2net} as the backbone to extract three feature maps. The feature map with the largest receptive field is first fed into the Augmented Shortcut Module (ASM). Subsequently, the feature map processed by ASM and the other two feature maps are input into the Efficient Hybrid Encoder (EHE). The ASM enhances feature representation and preserves details, while the EHE achieves better cross-scale fusion. Ultimately, the network integrates three detection heads to obtain detection results after classification and regression tasks.

\subsection{Augmented Shortcut Module} 
To address the issue of too single features extracted by the backbone, we propose a powerful Augmented Shortcut Module (ASM) to increase feature diversity in our EMDFNet through refining shortcut connections. 
ASM includes spatial feature selection, channel feature selection and Aug-S. The module parallelizes three branches and then outputs diversified features.
The final structure of the ASM is shown in Fig.~\ref{figure2}. The ASM can be formulated as follows:

\begin{equation}
\begin{aligned}
ASM = MLCA({\Gamma} (Z_{l},{\Theta}_{l} ),\ shortcut(Z_{l}),\  FS(Z_{l}))
\end{aligned}
\end{equation}

The augmented shortcut connection, denoted as $\Gamma$ with parameters $\Theta$, enhances the original shortcut by adding alternative paths that incorporate attention mechanisms. Unlike simple identity projection, $\Gamma$ transforms input features into a new space using various transformations dictated by $\Theta$. The feature selection (FS) utilizes Efficient Multi-scale Attention (EMA)\cite{ouyang2023efficient} to select spatial features, and all branches undergo processing by the Mixed Local Channel Attention (MLCA)\cite{wan2023mixed} channel selection module before concatenation.

\begin{figure}[!htbp]
	\centering
	\includegraphics[scale=0.7]{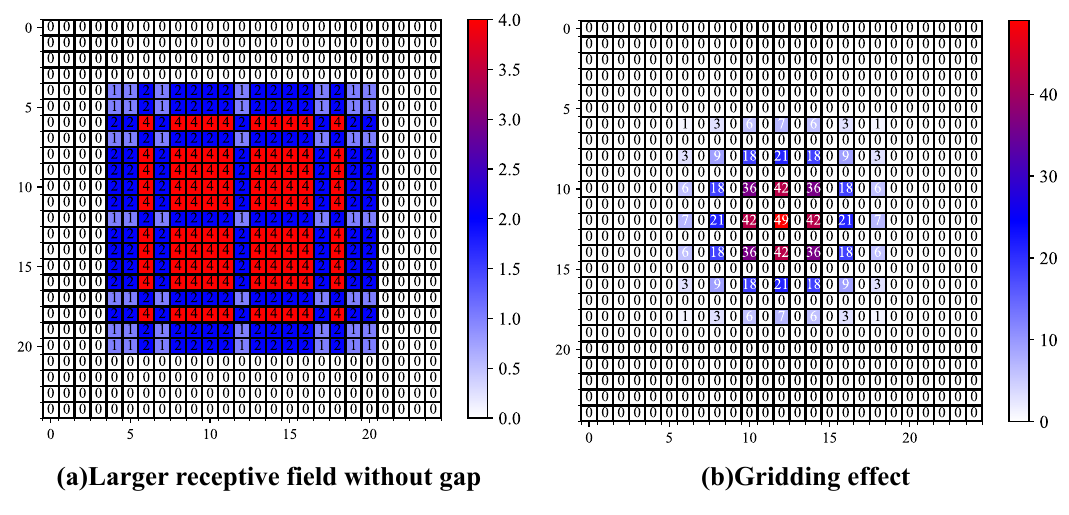}
	\caption{The dilation rates used in (a) are [1, 2, 5], where all pixel values are effectively utilized, while in (b), the dilation rates are [2, 2, 2], resulting in a gridding effect.}
\label{figure3}
\end{figure}

The original residual connection is updated with two main components: a feature selection layer for spatial and channel features, and an enhanced shortcut connection Aug-S.
We utilize spatial feature selection to uniformly distribute spatial semantics and utilize channel feature selection to fuse local and global channel information.
In addition, Aug-S consists of a set of dilated convolutions that expand the receptive field while keeping the image size unchanged. Aug-S enhances feature diversity as an alternative route. To preserve details lost during downsampling by the backbone, we utilize Hybrid Dilated Convolution (HDC) with dilation rates of $\left[1, 2, 5\right]$. This approach helps avoid the gridding effect associated with uniform dilation rates and ensuring comprehensive pixel utilization. As shown in Fig.~\ref{figure3}. This approach improves the retention of feature map details.

\vspace{-0.5cm}

\subsection{Efficient Hybrid Encoder}

To enhance the multi-scale fusion capability of the proposed EMDFNet, we introduce the Efficient Hybrid Encoder (EHE) for the first time in the field of traffic sign detection, as illustrated in Fig. \ref{figure2}. EHE is an excellent component accepted in CVPR 2024, which can achieve multi-scale fusion through intra-scale interaction and cross-scale interaction \cite{lv2023detrs}. EHE comprises two main units: an Attention-based Intrascale Feature Interaction (AIFI) for reducing computational redundancy by emphasizing high-level intra-scale interactions, and a CNN-based Cross-scale Feature-fusion Module (CCFM), which introduces fusion blocks to combine adjacent feature levels. As shown in formula \ref{flatten}, AIFI prioritizes high-level features for their richer semantic content, avoiding lower-level interactions that could lead to duplication or confusion. As shown in formula \ref{CCFM}, CCFM's fusion blocks, combine features through element-wise addition, optimizing feature integration to enhance object detection and recognition performance. 

\begin{equation}
\begin{aligned}
\label{flatten}
{F_{5}} &= \operatorname { AIFI }\left(S_{5}\right) 
\end{aligned}
\end{equation}
\begin{equation}
\begin{aligned}
\label{CCFM}
\text { Output } &= \operatorname{CCFM}\left(\left\{S_{3}, S_{4}, F_{5}\right\}\right)
\end{aligned}
\end{equation}
where $S3$, $S4$, $S5$, $F5$ are feature maps. AIFI first decomposes the feature map into sequence vectors, and then resynthesizes the feature map through the attention mechanism. CCFM fuses feature maps of different sizes.

\subsection{Loss Function}
IoU Loss has several limitations, including its inability to optimize when boxes do not intersect (IoU=0), its failure to indicate the distance between boxes, and its inaccuracy in showing the degree of overlap. To overcome IoU Loss limitations, we introduce SIoU Loss \cite{gevorgyan2022siou} for calculating rectangular box loss, enhancing training speed and detection precision. 




The SIoU Loss function is defined as follows:
\begin{equation}
\begin{aligned} 
L_{box} &= 1 - IoU + \frac{L_{dis} +L_{shape}}{2}
\end{aligned}
\end{equation}
\begin{equation}
\begin{aligned} 
L &= W_{box}L_{box}+W_{cls}L_{cls} 
\end{aligned}
\end{equation}
where $L_{box}$ represents shape cost. $IoU$ (Intersection over Union) represents the intersection of the union loss. $L_{dis}$ and $L_{shape}$ represent the distance cost and the shape loss, respectively. $W_{box}$ and $W_{cls}$ represent weights assigned to the bounding box and classification losses, respectively. $L_{cls}$ represents the focal loss.  From these we can obtain the final loss function $L$.

\section{Experiment} 

\subsection{Dataset} 

We use the TT100K and GTSDB datasets to evaluate our EMDFNet.

\begin{figure}[!htbp]
	\centering
	\includegraphics[scale=0.25]{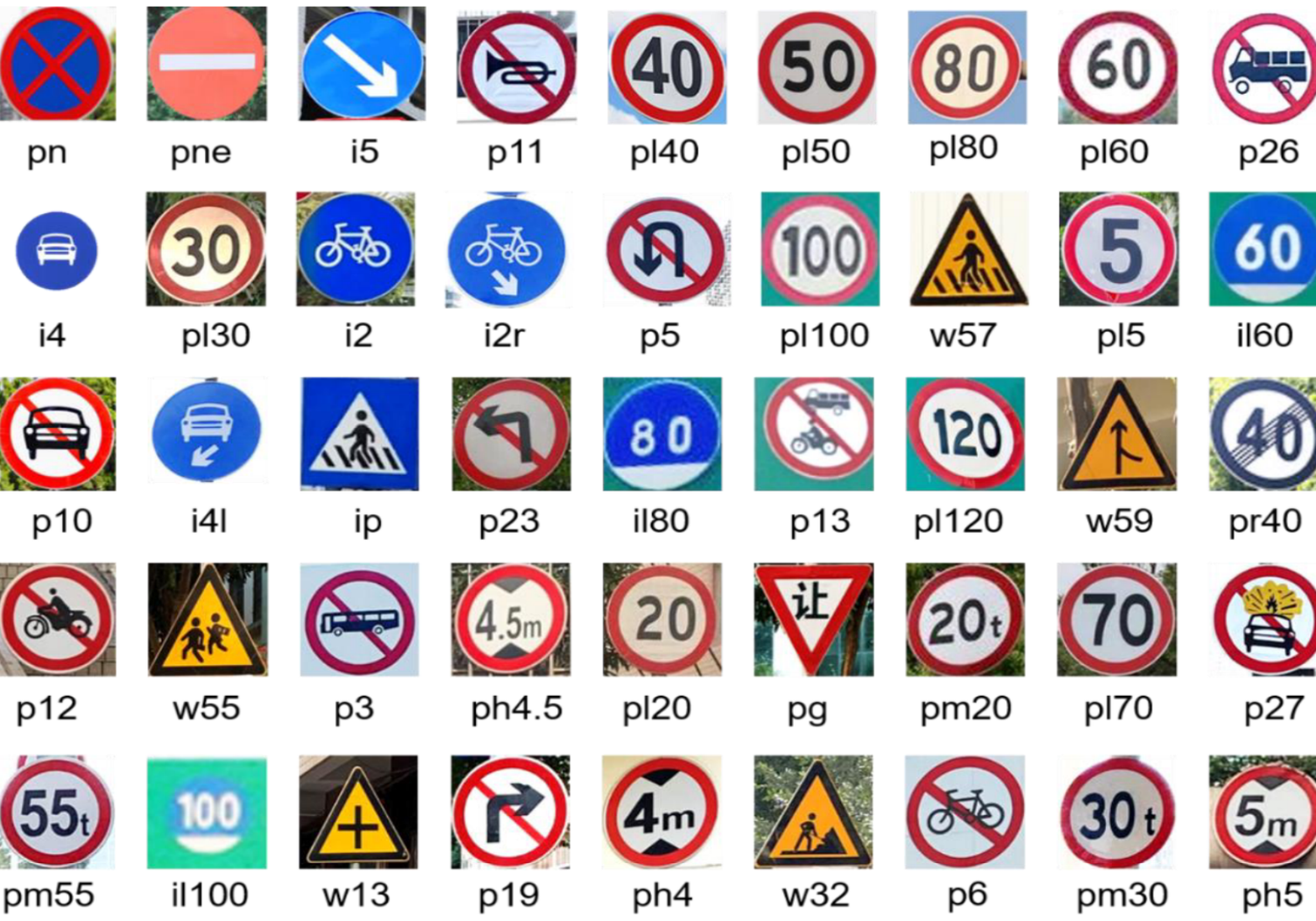}
	\caption{Illustrations of 45 remaining traffic sign categories from the TT100K dataset.}
\label{TT100K}
\end{figure}

\noindent 
$\bullet$ \textbf{TT100K} \cite{zhu2016traffic} provides 1,000,000 images and 30,000 annotated traffic signs that are collected from Tencent Street Views in Chinese Cities. The resolution of the provided images is uniform $2048 \times 2048$ pixels. We ignore categories with fewer than 100 instances following, thus leaving 45 categories to detect, as illustrated in Fig.~\ref{TT100K}. The publicly available benchmark dataset can be found at \href{http://cg.cs.tsinghua.edu.cn/traffic-sign/.}{http://cg.cs.tsinghua.edu.cn/traffic-sign/.} To avoid the impact of various augmentation policies, we fine-tune the proposed EMDFNet and other detectors on the original training set, which includes 6105 images, and test them on the original testing set, which comprises 3071 images, for a fair comparison. Three categories labeled as `ph5', `w32', and `wo' are excluded due to their mixed content of several traffic signs.

\noindent 
$\bullet$ \textbf{GTSDB}\cite{Houben-IJCNN-2013} provides 900 images and 1213 annotated traffic signs.
All these images are captured from video sequences recorded near Bochum, Germany. The benchmark dataset is publicly available at \href{https://benchmark.ini.rub.de/gtsdb_dataset.html}{https://benchmark.ini.rub.de/gtsdb\_dataset.html}. The final images are clipped to $1360 \times 800$ pixels. The standard task for GTSDB focuses on traffic signs, and all traffic signs are classified into four major categories: prohibitory, danger, mandatory, and other. 

\subsection{Implementation Details}  

We develop EMDFNet using PyTorch 1.8.0 on an RTX 3090 GPU, utilizing pretrained Res2Net-101 weights to expedite training.
Our batch size is 8 for $640\times640$ inputs, dropping to 4 for $1024\times1024$ inputs due to memory constraints. We apply multi-scale inputs, mosaics, and mixup augmentations for better generalization. Training involves 400 epochs on the TT100K dataset and 100 epochs on the GTSDB dataset, starting with a five-epoch warm-up at a zero learning rate. We utilize Stochastic Gradient Descent (SGD) with a weight decay of 0.0005 and momentum of 0.9. The learning rate is adjusted using linear warm-up and cosine annealing techniques, customizing for each stage of training.

\subsection{Evaluation Metrics}  

To assess network performance, we utilize Average Precision (AP), parameters, GFLOPs, and Frames Per Second (FPS) as metrics.





Average Precision (AP) is an evaluation index of detection accuracy, calculated as the average value of the average accuracy across multiple categories. AP is defined as the area under the curve, which depends on recall and accuracy. The function to calculate AP is shown below formula
 \ref{AP}. $P (R)$ is the curve based on the recall rate and precision rate. Mean Average Precision (mAP) is a comprehensive criterion that is shown below formula~\ref{mAP}. $N$ is the number of categories in the function. 
 
\begin{equation}
    \mathrm{AP}=\int_0^1 P(R) d R \\
    \label{AP}
\end{equation}

 \begin{equation} \text{mAP}@IoU = \frac{1}{N} \sum_{i=1}^{N} AP@IoU \label{mAP}\end{equation}

mAP@.5 represents the model's accuracy in object detection with a minimum 50\% overlap between predicted and actual bounding boxes. mAP@.75 evaluates precision with a stricter 75\% overlap requirement. mAP@.5:.95 averages mAP over $IoU$ thresholds from 0.5 to 0.95. It provides a comprehensive overview of the model's performance across a range of criteria from lenient to strict. In object detection, $AP_{s}$, $AP_{m}$, and $AP_{l}$ represent average precision metrics calculated for objects of small, medium, and large sizes detected within images.

Parameters calculate the number of parameter in the model, indicating memory resource usage. GFLOPs measure computational complexity as billions of operations per second. FPS, the rate of image processing, reflects real-time performance and is affected by image resolution. Higher resolutions reduce FPS with the same model and conditions.

\subsection{Quantitative Analysis}  
\subsubsection{Comparisons of SOTA on TT100K Dataset.}  
We compare the proposed approach with state-of-the-art traffic sign detection methods. Table \ref{tab:4} presents the mAP@.5, mAP@.75, mAP@.5:.95, $AP_{s}$, $AP_{m}$, and $AP_{l}$ values of the proposed method compared to other well-known detectors on the TT100K testing dataset. It can be observed that EMDFNet (640) and EMDFNet (1024) achieve 84.4\% mAP@.5 and 93.3\% mAP@.5, respectively.

From Table \ref{tab:4}, we can know when input sizes are similar, EMDFNet significantly outperforms other models. At an input size of 640, it achieves 84.4\%, surpassing SSD (68.7\%), ScratchDet (74.0\%), CAB Net (78.0\%), and Open-TransMind (83.0\%). At an input size of 1024, EMDFNet achieves 93.3\% accuracy. A 23.8\% increase over Sparse R-CNN's 69.5\%,  surpassing Mask R-CNN (70.8\%), Faster R-CNN (74.1\%), and Cascade R-CNN (80.1\%), setting a new benchmark in object detection.

EMDFNet outperforms models with larger inputs, reaching 84.4\% mAP@.5 at an input size of 640. This surpasses RetinaNet (61.9\%, 2048 input), Sparse R-CNN (69.5\%, 1024 input), Faster R-CNN (74.1\%, 1024 input), and Cascade R-CNN (80.1\%, 1024 input). Despite using smaller inputs, EMDFNet significantly enhances mAP, demonstrating its exceptional efficiency.


It can be seen from Table \ref{tab:4} that our network focuses on detecting small and medium traffic signs, with a particular emphasis on $AP_{s}$. Sparse R-CNN, Faster R-CNN, Cascade R-CNN, and SwinT + Cascade R-CNN achieve $AP_{s}$ of 35.6\%, 27.7\%, 36.2\%, and 44.5\%, respectively. EMDFNet achieves 55.4\% and 61.1\% $AP_{s}$ at input sizes of 640 and 1024, respectively, demonstrating its effectiveness in detecting smaller objects.

To further verify the performance of the proposed network, Table \ref{each category} shows the detection accuracy of EMDFNet in each category and compares it with other networks. From the Table \ref{each category} we can see that EMDFNet achieves SOTA in almost all categories. In particular, the accuracy of ‘p27’ reach 100.0\%, ‘il60’ reach 99.8\%, and ‘il80’ reach 98.9\%.

In summary, it can be concluded that our EMDFNet's efficacy is clear: Res2Net enhances feature extraction, Augmented Shortcut Module diversifies input features, and Efficient Hybrid Encoder enhances multi-scale integration, thereby enhancing small traffic sign detection. Additionally, larger input sizes yield higher $AP_{s}$, indicating the potential for further enhancements in small object detection performance.

\begin{table}[!t]
    \centering
    \caption{The performance of different models on the TT100K dataset. The best results are shown in bold. 
}
    \label{tab:4} 
    \resizebox{\linewidth}{!}{
    \begin{tabular}{l| m{1cm}<{\centering} | m{1.8cm}<{\centering} |m{1.5cm}<{\centering} m{1.6cm}<{\centering} m{1.7cm}<{\centering}| lll | m{1cm}<{\centering}m{1cm}<{\centering}m{1cm}<{\centering}ll}
    \toprule 
        Model & Year & Input Size & mAP@.5 & mAP@.75 & mAP@.5:.95 & $AP_{s}$ & $AP_{m}$ & $AP_{l}$ & FPS & Param. & GFLOPS &  \\ 
        \hline
        Faster R-CNN\cite{ren2015faster} & 2015 & 1024 $\times$ 1024 & 74.1 & 69.3 & 58.9 & 27.7 & 73.1 & 81.8 & 21 & 41.6 & 211.5  \\
        SSD512\cite{liu2016ssd} & 2016 & 512 $\times$ 512 & 68.7 & - & - & - & - & - & - & - & - \\
        TT100K\cite{zhu2016traffic} & 2016 & 640 $\times$ 640 & 83.3 & 79.0 & 65.8 & - &  - & - & - & 35.1 & - \\
        RetinaNet\cite{lin2017focal} & 2017 & 2048 $\times$ 2048 & 61.9 & - & - & - & - & - & - & - & - \\ 
        Mask R-CNN\cite{he2017mask} & 2017 & 1000 $\times$ 800 & 70.8 & - & - & - & - & - & - & - & - \\ 
        Cascade-RCNN\cite{cai2018cascade} & 2018 & 1024 $\times$ 1024 & 80.1 & 74.8 & 63.5 & 36.2 & 74.6 & 83.0 & 17.0 & 69.3 & 239.2  \\
        ScratchDet\cite{zhu2019scratchdet} & 2019 & 512 $\times$ 512 & 74.0  & - & - & - & - & - & - & - & - \\
        FCOS\cite{tian2019fcos} & 2019 & 2048 $\times$ 2048 & 83.3 & - & - & - & - & - & - & - & - \\ 
        PCN\cite{liang2020traffic} & 2020 & 2048 $\times$ 2048 & 92.0 & 86.6 & 72.1 & - & - & - & - & 34.3 & - \\ 
        Sparse R-CNN\cite{sun2021sparse} & 2021 & 1024 $\times$ 1024 & 69.5 & 62.8 & 53.2 & 35.6 & 61.2 & 70.6 & 21.0 & 106.2 & 153.3  \\ 
        Swint-T+Cascade R-CNN\cite{liu2021swin} & 2021 & 1024 $\times$ 1024 & 86.3 & 79.1 & 68.6 & 44.5 & 73.7 & 85.2 & 14.0 & 74.6 & 245.5  \\ 
        CAB Net\cite{cui2020context} & 2022 & 512 $\times$ 512 & 78.0  & - & - & - & - & - & - & - & - \\ 
        TRD-YOLO\cite{s23083871} & 2023 & 640 $\times$ 640 & 37 .5 & - & 30.1 & 26.8 & 34.5 & 40.2 & - & - & -  \\
        YOLO-SG\cite{han2023yolo} & 2023 & 640 $\times$ 640 & 75.8 & - & - & - & - & 60.2 & - & -  \\
        Open-TransMind\cite{shi2023open} & 2023 & 512 $\times$ 512 & 83.0 & - & - & - & - & - & - & - & - \\
        Transformer Fusion\cite{10422764} & 2024 & 640 $\times$ 640 & 53.5 & - & 39.5 & 43.6 & 43.7 & 58.6 & - & - & -  \\\hline
        \textbf{EMDFNet(ours)} & - & 640 $\times$ 640 & 84.4 & 76.8 & 64.9 & 55.4 & 68.5 & 81.7 & \textbf{46.7} & 29.7 & \textbf{98.4}  \\ 
        \textbf{EMDFNet(ours)} & - & 1024 $\times$ 1024 & \textbf{93.3} & \textbf{87.1} & \textbf{73.8} & \textbf{61.1} & \textbf{78.8} & \textbf{85.4} & 31.0 & \textbf{29.6} & 187.8  \\
        \bottomrule 
    \end{tabular}
    }
\end{table}

\begin{table}[!ht]
    \centering
    \caption{Comparisons of AP for each category on the TT100K testing set. Each column represents a traffic sign. The EMDFNet outperforms other detectors, achieving the SOTA performance.}
    \label{each category}
    \resizebox{\linewidth}{!}{
    \begin{tabular}{l|llllllllllllll}
    \toprule 
        Method & i2 & i4 & i5 & il100 & il60 & il80 & io & ip & p10 & p11 & p12 & p19 & p23 & p26 \\ \hline
        Faster R-CNN\cite{ren2015faster} & 44.0& 46.0& 45.0& 41.0& 57.0& 62.0& 41.0& 39.0& 45.0& 38.0& 60.0& 59.0& 65.0& 50.0\\ 
        Faster R-CNN\cite{he2016deep} & 59.3 & 73.8 & 79.7 & 76.6 & 76.3 & 68.5 & 64.9 & 66.8 & 52.2 & 58.5 & 45.9 & 48.2 & 74.4 & 66.1 \\ 
        FPN\cite{lin2017feature} & 72.5 & 79.6 & 88.3 & 90.2 & 88.2 & 84.9 & 77.4 & 75.8 & 62.7 & 75.9 & 60.2 & 53.7 & 75.8 & 76.0 \\ 
        Mask R-CNN\cite{he2017mask} & 71.4 & 85.6 & 89.0 & 89.4 & 86.3 & 82.3 & 78.0 & 77.6 & 59.6 & 76.9 & 63.8 & 52.0 & 72.9 & 81.7 \\ \hline
        SSD512\cite{liu2016ssd} & 70.1 & 79.3 & 85.3 & 77.1 & 86.4 & 78.7 & 72.3 & 71.6 & 64.5 & 57.1 & 67.7 & 73.0 & 80.4 & 70.7 \\ 
        DSSD512\cite{fu2017dssd} & 65.0  & 86.2 & 88.6 & 62.7 & 87.7 & 76.2 & 60.2 & 85.5 & 66.2 & 55.1 & 54.4 & 78.4 & 79.3 & 75.5 \\ 
        RFB Net 512\cite{liu2018receptive} & 75.6 & 79.4 & 87.9 & 87.4 & 89.9 & 88.4 & 77.2 & 79.0  & 66.1 & 66.9 & 71.1 & 72.8 & 83.4 & 74.9 \\ 
        ScratchDet\cite{zhu2019scratchdet} & 76.6 & 86.9 & 89.2 & 82.2 & 88.8 & 81.3 & 73.9 & 77.3 & 68.8 & 65.3 & 70.8 & 67.2 & 80.2 & 74.9 \\ 
        CAB Net\cite{cui2020context} & 76.0  & 87.5 & 89.4 & 80.6 & 89.9 & 85.3 & 80.5 & 78.0  & 69.1 & 77.6 & 74.3 & 87.6 & 87.1 & 81.4 \\ 
        CAB-s Net\cite{cui2020context} & 75.2 & 86.4 & 89.4 & 84.9 & 89.2 & 89.1 & 81.6 & 77.8 & 69.7 & 72.4 & 72.3 & \textbf{89.0}  & 88.3 & 81.6 \\ \hline
        \textbf{EMDFNet(ours)} & \textbf{84.0}  & \textbf{93.5}  & \textbf{96.1}  & \textbf{95.1}  & \textbf{99.8}  & \textbf{98.9}  & \textbf{84.6}  & \textbf{86.9}  & \textbf{81.8}  & \textbf{89.9}  & \textbf{85.8}  & 81.5  & \textbf{89.9}  & \textbf{87.6}  \\ \bottomrule 
        \toprule 
        Method & p3 & p5 & p6 & pg & ph4 & ph4.5 & pl100 & pl120 & pl20 & pl30 & pl40 & pl5 & pl50 & pl60 \\ \hline
        Faster R-CNN\cite{ren2015faster} & 48.0& 57.0& 75.0& 80.0& 68.0& 58.0& 68.0& 67.0& 51.0& 43.0& 52.0& 53.0& 39.0& 53.0\\ 
        Faster R-CNN\cite{he2016deep} & 65.4 & 74.9 & 39.1 & 78.2 & 58.0 & 36.5 & 77.6 & 74.6 & 40.5 & 48.5 & 60.2 & 65.4 & 49.0 & 51.2 \\ 
        FPN\cite{lin2017feature} & 71.6 & 79.2 & 39.1 & 78.2 & 58.0 & 36.5 & 87.5 & 85.5 & 55.7 & 55.6 & 71.5 & 77.3 & 60.8 & 58.7 \\ 
        Mask R-CNN\cite{he2017mask} & 78.5 & 78.9 & 48.3 & 88.5 & 63.9 & 58.1 & 86.7 & 82.4 & 58.6 & 53.3 & 68.2 & 76.4 & 63.5 & 56.6 \\ \hline
        SSD512\cite{liu2016ssd} & 66.5 & 74.9 & 63.9 & 84.2 & 62.1 & 51.2 & 85.1 & 84.2 & 45.4 & 66.6 & 65.7 & 60.5 & 58.3 & 64.0 \\
        DSSD512\cite{fu2017dssd} & 56.1 & 79.6 & 55.4 & 85.8 & 60.7 & 88.6 & 79.1 & 69.6 & 65.3 & 68.3 & 68.2 & 61.5 & 65.5 & 64.7 \\
        RFB Net 512\cite{liu2018receptive} & 69.0  & 77.6 & 68.8 & 88.9 & 67.6 & 63.0  & 88.8 & 84.9 & 66.8 & 71.8 & 71.6 & 75.0  & 62.9 & 70.4 \\
        ScratchDet\cite{zhu2019scratchdet} & 71.2 & 87.3 & 65.4 & 79.1 & 66.8 & 55.7 & 85.8 & 84.7 & 63.6 & 67.1 & 73.2 & 65.4 & 69.9 & 72.8 \\
        CAB Net\cite{cui2020context} & 74.7 & 84.5 & 82.5 & 87.5 & \textbf{71.8} & 64.4 & 88.4 & 87.9 & 68.6 & 73.3 & 74.8 & 79.3 & 75.1 & \textbf{76.3} \\
        CAB-s Net\cite{cui2020context} & 76.8 & 85.4 & 78.0  & 86.4 & 71.7 & 62.3 & 89.2 & 88.7 & 71.5 & 73.5 & 75.3 & 75.9 & 73.1 & 75.9 \\ \hline
        \textbf{EMDFNet(ours)} & \textbf{88.0}  & \textbf{93.6}  & \textbf{82.9}  & \textbf{94.9}  & 63.2  & \textbf{88.8 } & \textbf{92.5}  & \textbf{91.6}  & \textbf{78.0}  & \textbf{80.0}  & \textbf{85.0}  & \textbf{89.2}  & \textbf{84.1}  & 75.3  \\ \bottomrule 
        \toprule 
        Method & pl80 & pm20 & pm30 & pm55 & pn & pne & po & pr40 & w13 & w55 & w57 & w59 & p27 & pl70 \\ \hline
        Faster R-CNN\cite{ren2015faster} & 52.0& 61.0& 67.0& 61.0 & 37.0 & 47.0 & 37.0 & 75.0 & 33.0 & 39.0 & 48.0 & 39.0 & 79.0 & 61.0 \\
        Faster R-CNN\cite{he2016deep} & 59.0 & 50.5 & 29.1 & 68.5 & 77.8 & 87.5 & 47.7 & 86.9 & 30.9 & 62.1 & 67.0 & 57.2 & 64.3 & 61.2 \\ 
        FPN\cite{lin2017feature} & 70.9 & 55.5 & 40.1 & 75.7 & 89.0 & 89.8 & 60.2 & 87.6 & 45.3 & 65.9 & 70.3 & 62.3 & 84.8 & 63.5 \\ 
        Mask R-CNN\cite{he2017mask} & 71.5 & 58.0 & 41.5 & 68.8 & 88.6 & 90.5 & 63.0 & 87.5 & 51.3 & 66.6 & 71.1 & 61.8 & 87.5 & 66.3 \\ \hline
        SSD512\cite{liu2016ssd} & 70.5 & 69.6 & 51.3 & 71.2 & 71.7 & 86.4 & 51.8 & 87.9 & 46.1 & 64.6 & 74.0 & 58.8 & 76.2 & 70.6 \\ 
        DSSD512\cite{fu2017dssd} & 75.6 & 66.3 & 50.6 & 76.5 & 67.2 & 88.9 & 51.7 & 88.0  & 60.6 & 70.1 & 83.6 & 75.1 & 60.9 & 71.0  \\ 
        RFB Net 512\cite{liu2018receptive}  & 71.9 & 73.7 & 54.0  & \textbf{86.5} & 78.0& 88.2 & 59.8 & 84.5 & 64.8 & 72.4 & 81.5 & 69.3 & 79.8 & 64.9 \\ 
        ScratchDet\cite{zhu2019scratchdet} & 75.9 & 73.3 & 52.2 & 76.5 & 76.7 & 89.4 & 62.9 & 85.0  & 69.1 & 70.3 & 84.7 & 76.5 & 79.7 & 70.2 \\ 
        CAB Net\cite{cui2020context} & 78.8 & 73.8 & \textbf{67.3} & 80.5 & 85.4 & 89.5 & 63.5 & 88.9 & 70.7 & 66.8 & 83.5 & 79.4 & 81.0  & \textbf{72.9} \\ 
        CAB-s Net\cite{cui2020context} & 78.4 & 71.2 & 67.0  & 83.3 & 82.2 & 89.2 & 64.0  & 88.2 & 57.2 & 75.2 & 80.1 & 66.6 & 87.5 & 70.7 \\ \hline
        \textbf{EMDFNet(ours)} & \textbf{80.5}  & \textbf{79.5}  & 60.3  & 83.5  & \textbf{96.3}  & \textbf{96.6}  & \textbf{80.7}  & \textbf{92.7}  & \textbf{92.7}  & \textbf{80.9}  & \textbf{87.8}  & \textbf{87.9}  & \textbf{100.0}  & 72.0 \\
        \bottomrule 
    \end{tabular}
    }
\end{table}

\subsubsection{Comparisons of SOTA on GTSDB Dataset.}  

To further verify the performance of our EMDFNet, we conduct experiments based on the GTSDB dataset. We fix the input size of the model to match the original image size of the dataset for comparison under consistent condition. The experimental results are shown in Table \ref{GTSDB com}. From Table \ref{GTSDB com}, we can observe that under the same input condition, the mAP@.5 value of EMDFNet reach 97.0\%, significantly surpassing the performance of other models. Among them, the most notable improvement is that EMDFNet is 35.4\% higher than SSD Mobilenet (from 61.6\% to 97.0\%).

When compared with Faster R-CNN Inception ResNet V2, the increase in mAP@.5 value is significant (from 95.7\% to 97.0\%), and the FPS of EMDFNet is about eight times higher, with a significant reduction in the number of parameters and computational cost. This demonstrates that EMDFNet not only performs well in accuracy but also achieves good results in speed and real-time performance.

Through comparisons of different datasets (TT100K and GTSDB), it is found that EMDFNet consistently achieves state-of-the-art performance levels, indicating that EMDFNet demonstrates excellent generalization capability.

\vspace{-0.5cm}

\subsubsection{Analyzing Object Detection Errors.} 
We compare EMDFNet and YOLOX using the TIDE toolkit to tabulate error distributions on the TT100K testing set, as presented in Table \ref{tide}. EMDFNet exhibites a lower classification error than YOLOX (10.02\% vs. 11.22\%), attributed to the Augmented Shortcut Module enhancing feature diversity and Efficient Hybrid Encoder improving feature spatial distribution for classification. EMDFNet also demonstrates a decreased miss error rate (0.98\% vs. 1.87\%), attributed to its effective feature extraction, which reduces the omission of essential details, thereby enhancing localization and classification accuracy.  Both and Dupe are basically the same, which may be caused by dataset noise. The remaining error rates decrease.

\begin{table}[!ht]
    \centering
    \caption{The performance of different models on the GTSDB dataset.}
    \begin{tabular}{l|m{1.5cm}<{\centering}|m{1cm}<{\centering} m{1cm}<{\centering} m{1cm}<{\centering}}
    \toprule 
        Model & mAP@.5 & FPS & Params & GFLOPS \\ \hline
        SSD Mobilenet\cite{liu2016ssd} & 61.6 & 66.0 & 5.5 & 2.3  \\ 
        SSD Inception V2\cite{liu2016ssd} & 66.1  & 42.1 & 13.4 & 7.5 \\ 
        Cascade R-CNN\cite{cai2018cascade} & 70.4 & - & - & - \\ 
        Faster R-CNN + IFA-FPN\cite{tang2021integrated} & 78.0 & - & - & - \\ 
        YOLO V2\cite{redmon2017yolo9000} & 78.8 & 46.5 & 50.5 & 62.7 \\ 
        Cascade R-CNN + IFA-FPN\cite{tang2021integrated} & 80.3 & - & - & - \\ 
        Faster R-CNN Inception V2\cite{ren2015faster} & 90.6 & 17.0 & 12.8 & 120.6 \\ 
        Faster R-CNN ResNet 50\cite{ren2015faster} & 91.5 & 9.6 & 43.3 & 533.5 \\ 
        Faster R-CNN ResNet 101\cite{ren2015faster} & 95.0 & 8.1 & 62.3 & 625.7 \\ 
        R-FCN ResNet 101 & 95.1 & 11.7 & 64.5 & 269.9 \\ 
        Faster R-CNN Inception ResNet V2\cite{ren2015faster} & 95.7 & 2.2 & 59.4 & 1837.5 \\ \hline
        \textbf{EMDFNet(ours)} & \textbf{97.0} & 16.7 & 29.7 & 190.4 \\ 
        \bottomrule 
    \end{tabular}
    \label{GTSDB com}
\end{table}

\vspace{-0.2cm}

\subsection{Qualitative Analysis} 
\begin{figure}[H]
	\centering
        \setlength{\abovecaptionskip}{0.cm}
	\includegraphics[scale=0.39]{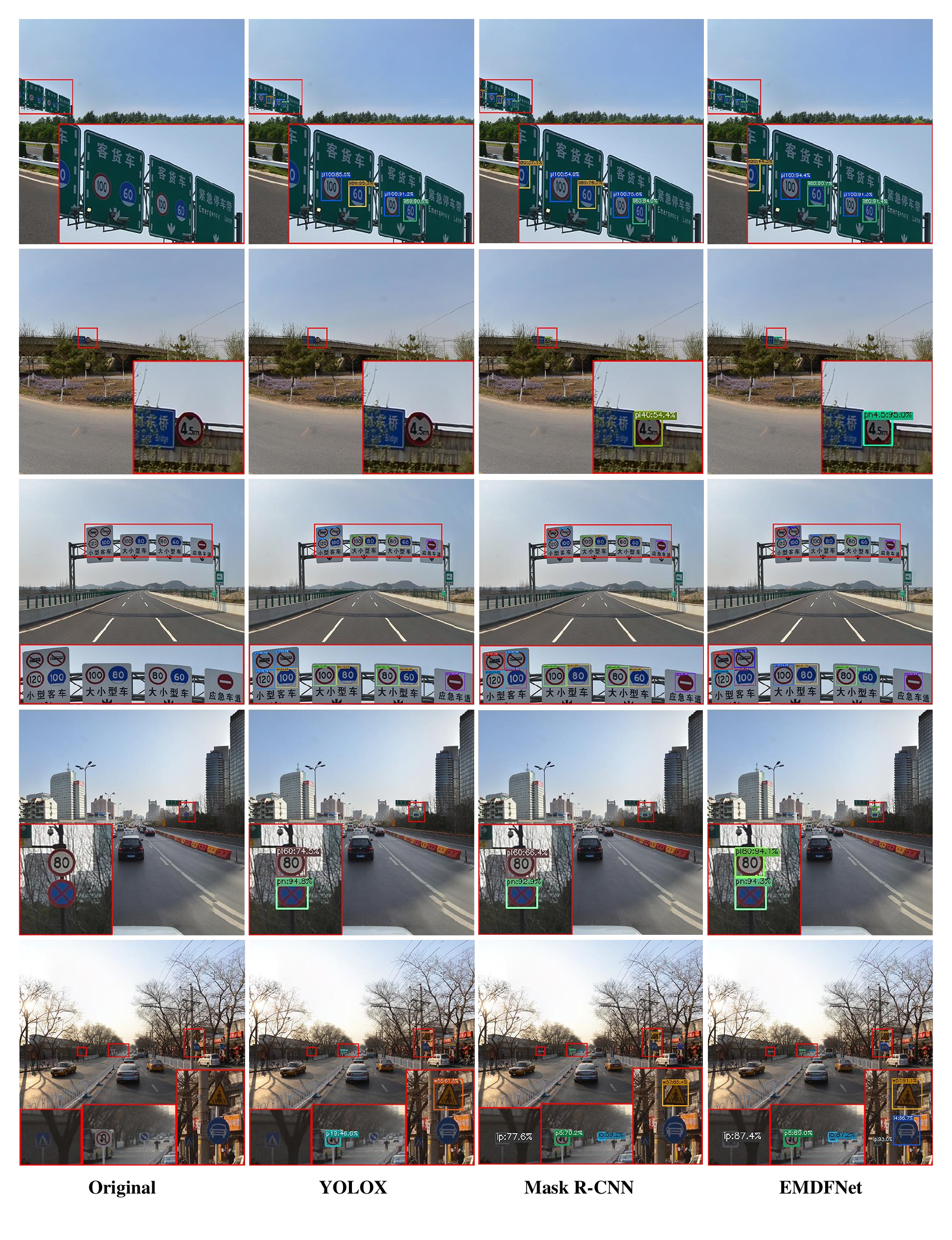}
	\caption{Comparison of detection performance between EMDFNet and other models on the TT100K testing set. Zoom in to see details.}
\label{vis result}
\end{figure}

To demonstrate the effectiveness of EMDFNet in small object detection, we conduct visualization experiments on the TT100K testing set, as illustrated in Fig. \ref{vis result}.

As shown in the second line of the Fig.\ref{vis result}, there are a total of six ground truth objects in the images. MDNet successfully detects all six objects and correctly classifies them with high confidence scores. This can be attributed to the good performance of MDNet in detecting small objects, demonstrating its robustness and high stability in small traffic sign detection. In contrast, YOLOX only detects two of them and Mask R-CNN only detects four of them. More dramatically, the detected objects `p19' and `w55' by YOLOX are completely different from the ground truth objects `p5' and `w57'. This discrepancy may be attributed to that the objects are relatively distant and small in the images, resulting in poor classification and detection performance.
Results framed in red highlight EMDFNet's ability to accurately detect and classify small, distant, or edge-located signs with high confidence, addressing YOLOX and Mask R-CNN limitations in edge detection and small object classification.

\subsection{Ablation Study} 
To investigate the performance of various components of our EMDFNet, we add the separate component to the baseline to observe the detection performance. All experiments are performed on the TT100K dataset.
The ablation results about ASM, EHE, SIoU and Res2net are illustrated in Table \ref{tab:2}.

\noindent 
\textbf{Impact of Augmented Shortcut Module.} 
we first verify ASM's feature diversity capability. As observed in Table \ref{tab:2}, mAP@.5 increase by 0.9\%, mAP@.75 increase by 1.5\%, and mAP@.5:.95 increase by 1.4\%. The results show that the mAP value comprehensively improves, proving the ability of ASM feature diversity. ASM enhances input feature diversity beyond standard shortcuts by introducing a transformative multiple branches.

\noindent 
\textbf{Impact of Efficient Hybrid Encoder.} 
To preserve crucial discriminative information and enhance feature fusion during aggregation and path selection, we integrate EMDFNet's EHE into YOLOX, resulting in significant improvements across various metrics.
According to experimental in Table \ref{tab:2}, $AP_{s}$, $AP_{m}$, and $AP_{l}$ comprehensively enhance. The increase in $AP_{s}$, $AP_{m}$, and $AP_{l}$ are 2.0\%, 1.5\%, and 3.5\%, respectively. Therefore, EHE has a very good effect on the fusion of small, medium and large targets.

\noindent 
\textbf{Impact of SIoU. } 
To verify the impact of different loss functions, we conduct ablation experiments on the original IoU Loss and SIoU Loss. EMDFNet's SIoU Loss considers spatial distances and angles between predicted and actual bounding boxes, improving training efficiency compared to YOLOX's traditional IoU Loss.
Experimental results in Table \ref{tab:2} confirm these improvements, enhancing metrics such as mAP@.5, mAP@.75, and mAP@.5:.95, without adding complexity. Notably, mAP@.5 increases by 1.2\% to 81.8\%.

\begin{table*}[!h]
    \centering
    \caption{Detection error distribution of YOLOX and EMDFNet on the TT100K testing set. TIDE categorizes errors into six types: Cls (correct localization but incorrect classification), Loc (correct classification but incorrect localization), Both (incorrect classification and localization), Dupe detection (correct classification and localization with a prior matching detection), Bkg (detected background as foreground), and Miss (undetected ground truths outside of classification or localization errors).}
    \label{tide}  
    \begin{tabular}{l|llllll|m{2cm}<{\centering} m{2cm}<{\centering}}
    \toprule 
        Model & Cls & Loc & Both & Dupe & Bkg & Miss & FalsePos(FP) & FalseNeg(FN) \\ \hline
        YOLOX & 11.22 & 0.44 & \textbf{0.01} & \textbf{0.05} & 2.71 & 1.87 & 7.90 & 9.67 \\ 
        \textbf{EMDFNet(ours)} & \textbf{10.02} & \textbf{0.31} & 0.03 & 0.08 & \textbf{2.63} & \textbf{0.98} & \textbf{7.47} & \textbf{7.62} \\ 
    \bottomrule 
    \end{tabular}
\end{table*}

\begin{table}[!t]
	\centering
	\caption{Ablation Experiment on the TT100K Dataset. We conduct ablation experiments on ASM, EHE, SIoU  and Res2Net.}
	\label{tab:2}  

	\begin{tabular}{cccc|ccc}
		\toprule 
		Model & mAP@.5 & mAP@.75 & mAP@.5:.95 & $AP_{s}$ & $AP_{m}$ & $AP_{l}$\\
		\hline 
		YOLOX\cite{ge2021yolox} & 80.6 & 72.8 & 61.2 & 53.1 & 66.1 & 72.4 \\
		YOLOX+ASM  & 81.5 & 74.3 & 62.6 & 52.7 & 67.8 & 73.0 \\
		YOLOX+EHE  & 82.2 & 74.6 & 63.1 & 55.1 & 67.6 & 75.9 \\
  		YOLOX+SIoU  & 81.8 & 73.0 & 61.7 & 53.4 & 66.6 & 71.2 \\
        YOLOX+Res2Net  & 82.5 & 74.8 & 63.2 & 54.0 & 68.4 & 76.0 \\
        \hline
        \textbf{EMDFNet(ours)} & \textbf{84.4} & \textbf{76.8} & \textbf{64.9} & \textbf{55.4} & \textbf{68.5} & \textbf{81.7} \\
		\bottomrule 
	\end{tabular}
\end{table}

\noindent 
\textbf{Impact of Res2Net.} 
Finally, we assess EMDFNet's Res2Net backbone, which enhances multi-scale features and expands the receptive field of per layer compared to YOLOX's Darknet53. Theoretically, Res2Net offers superior feature extraction, leading to enhanced detection performance. By partitioning input features into different groups and processing them through varying numbers of convolutional blocks, Res2Net achieves multi-scale effect, which is particularly advantageous for small object detection. Experimental results in Table \ref{tab:2}    demonstrate improvements across all metrics, with mAP@.5, mAP@.75, mAP@.5:.95, $AP_{s}$, $AP_{m}$, and $AP_{l}$ increasing by 1.9\%, 2.0\%, 2.0\%, 0.9\%, 2.3\%, and 3.6\%, respectively. These enhancements, achieving with minimal increase in parameters and computational complexity, affirm the efficacy of Res2Net in EMDFNet.

\section{Conclusion}     
This paper proposes a novel detector EMDFNet to address the issues of feature singularity and weak multi-scale fusion capability. EMDFNet consists of two key components: an Augmented Shortcut Module and an Efficient Hybrid Encoder. The Augmented Shortcut Module utilizes extracted features as input for different branches, conducts various feature selection operations, and ultimately combines them to enhance feature diversity. The Efficient Hybrid Encoder performs information interaction based on intra-scale feature interaction and cross-scale feature fusion. Extensive experiments on challenging datasets such as TT100K and GTSDB demonstrate that our EMDFNet achieves state-of-the-art performance. We will continue to explore lightweight network to better apply them in autonomous driving.

{ 
\bibliographystyle{IEEEtran}
\bibliography{reference}
}
\end{document}